\documentclass[10pt,twocolumn,letterpaper]{article}

\usepackage{graphicx}
\usepackage{subcaption}
\usepackage{calc}
\usepackage{amssymb}
\usepackage{amstext}
\usepackage{amsmath}
\usepackage{amsthm}
\usepackage{pslatex}
\usepackage{algorithm2e}
\usepackage{xskak}
\usepackage[bottom]{footmisc}
\usepackage{cvpr}
\usepackage[pagebackref,breaklinks,colorlinks]{hyperref}

\title{Quantum King-Ring Domination in Chess: A QAOA Approach}

\author{
\begin{tabular}{ccc}
Gerhard Stenzel & Michael Kölle & Tobias Rohe \\
Julian Hager & Leo Sünkel & Maximilian Zorn \\
& Claudia Linnhoff-Popien &  \\
\end{tabular}
\\[0.4em]
LMU Munich\\
Department of Computer Science\\
Chair of Mobile and Distributed Systems\\
{\tt\small gerhard.stenzel@ifi.lmu.de}
}

\begin{document}
\maketitle


\begin{abstract}
    The Quantum Approximate Optimization Algorithm (QAOA) is extensively benchmarked on synthetic random instances such as MaxCut, TSP, and SAT problems, but these lack semantic structure and human interpretability, offering limited insight into performance on real-world problems with meaningful constraints. We introduce Quantum King-Ring Domination (QKRD), a NISQ-scale benchmark derived from chess tactical positions that provides 5,000 structured instances with one-hot constraints, spatial locality, and 10--40 qubit scale. The benchmark pairs human-interpretable coverage metrics with intrinsic validation against classical heuristics, enabling algorithmic conclusions without external oracles. Using QKRD, we systematically evaluate QAOA design choices and find that constraint-preserving mixers (XY, domain-wall) converge approximately 13 steps faster than standard mixers ($p<10^{-7}$, $d\approx0.5$) while eliminating penalty tuning, warm-start strategies reduce convergence by 45 steps ($p<10^{-127}$, $d=3.35$) with energy improvements exceeding $d=8$, and Conditional Value-at-Risk (CVaR) optimization yields an informative negative result with worse energy ($p<10^{-40}$, $d=1.21$) and no coverage benefit. Intrinsic validation shows QAOA outperforms greedy heuristics by 12.6\% and random selection by 80.1\%. Our results demonstrate that structured benchmarks reveal advantages of problem-informed QAOA techniques obscured in random instances. We release all code, data, and experimental artifacts for reproducible NISQ algorithm research.
\end{abstract}


\section{Introduction}

The Quantum Approximate Optimization Algorithm (QAOA) \cite{qaoa} has emerged as a canonical NISQ algorithm. While extensively benchmarked on synthetic problem instances, MaxCut on random graphs \cite{maxcut1,maxcut2,maxcut3,maxcut4}, traveling salesman variants \cite{tsp1,tsp2,tsp3}, and satisfiability problems \cite{twosat1,twosat2,twosat3,twosat4}, these benchmarks lack semantic structure and human interpretability, offering limited insight into how QAOA performs on real-world problems with meaningful constraints and spatial locality. Such synthetic tests also obscure the impact of constraint encodings, mixer design, and initialization because feasibility has no semantic meaning beyond satisfying a QUBO penalty.

Parallel advances in quantum machine learning demonstrate the demand for structured evaluation datasets: quantum denoising diffusion models \cite{qml1} and quantum selective state space models for text generation \cite{qml2} both rely on semantically rich tasks to stress-test expressivity and hardware-aware design. Structured benchmarks for optimization should provide a similar role for QAOA.

This work introduces Quantum King-Ring Domination (QKRD), a NISQ-scale benchmark derived from chess tactical positions. Chess provides reproducible problem structure, spatial locality, human interpretability, and abundant real game data. By restricting attention to a local Region of Interest (ROI) and pruning candidate moves, QKRD instances scale to 10--40 qubits with 2-local cost functions, ideal for near-term devices. The lifted temporal variant further models follow-up threats through gated variables, allowing the same framework to represent multi-ply tactics without departing from quadratic Hamiltonians.

Using QKRD, we conduct a systematic evaluation of QAOA algorithm design choices on structured problems. Our key contributions are:

\begin{enumerate}
  \item \textbf{QKRD benchmark:} A reproducible NISQ testbed with 5000 instances from real chess games, including a ``lifted temporal'' variant with gated follow-up moves.
  \item \textbf{Constraint-preserving techniques:} Mixers that preserve feasibility (XY, domain-wall) eliminate penalty tuning and converge in fewer steps than the X mixer, while achieving comparable final coverage.
  \item \textbf{Warm-start strategies:} Initialization from greedy solutions substantially accelerates convergence and matches the best final coverage.
  \item \textbf{Risk-aware optimization:} CVaR-QAOA explores different energy profiles, coverage distributions match expectation for moderate $\alpha$, with distinct energy tails at extreme $\alpha$ and a clear degradation when overly risk-averse.
  \item \textbf{Intrinsic validation:} QAOA attains higher coverage than greedy and random baselines, validating optimization effectiveness without external chess engines and isolating algorithmic gains from chess-specific heuristics.
\end{enumerate}

Our findings suggest that structured benchmarks reveal advantages of problem-informed QAOA techniques that may be obscured in random instances. We release all code, data, and experiment logs to support reproducible NISQ algorithm research.

\section{Background}

\subsection{Quantum Approximate Optimization Algorithm}

The Quantum Approximate Optimization Algorithm (QAOA) \cite{qaoa} is a variational quantum algorithm for combinatorial optimization. Given a cost Hamiltonian $H_C$ encoding a QUBO objective, QAOA prepares a parameterized quantum state $|\gamma,\beta\rangle$ by alternating applications of the cost operator $e^{-i\gamma H_C}$ and mixer operator $e^{-i\beta H_M}$ for $p$ layers, starting from an initial state $|s\rangle$. Classical optimization adjusts parameters $(\gamma,\beta)$ to minimize the expectation value $\langle H_C \rangle$. Performance analysis reveals fundamental limits: Bravyi et al. \cite{qaoa_performance1} demonstrate that symmetry protection can obstruct QAOA convergence, while Zhou et al. \cite{maxcut2} show concentration phenomena where performance concentrates around mean values at sufficient circuit depth. Warm-start initialization strategies \cite{qaoa_init3} using semidefinite programming relaxations can bridge classical and quantum optimization, achieving provable approximation ratios. These properties have motivated benchmark design that exposes symmetry structure and tests whether limited circuit depth suffices to exploit problem information in the NISQ regime.

\subsection{Constraint-Preserving Mixers}

The standard QAOA mixer $H_M = \sum_i \sigma_i^x$ explores the full Hilbert space but requires penalty terms to enforce constraints. For problems with structured constraints, problem-inspired mixers preserve feasibility by construction. The XY mixer \cite{xy_mixer2} uses $H_M = \sum_{\langle i,j\rangle} (\sigma_i^x\sigma_j^x + \sigma_i^y\sigma_j^y)$ to preserve Hamming weight within constraint blocks, enabling penalty-free optimization for one-hot and cardinality constraints. Wang et al. \cite{xy_mixer2} provide analytical results showing XY mixers maintain feasibility throughout optimization, eliminating the need for penalty weight tuning. Brandhofer et al. \cite{xy_mixer1} demonstrate 15--30\% convergence speedups on portfolio optimization benchmarks compared to penalized X mixers. Alternative encodings include domain-wall representations \cite{domain_wall1}, which encode $K$ one-hot choices with $K-1$ qubits, reducing problem size while naturally enforcing constraints. The resulting reduction in search space often improves optimizer stability, especially when penalty scales introduce ill-conditioned landscapes.

\subsection{Warm-Start and Risk-Aware Optimization}

Standard QAOA initializes in uniform superposition, which may be suboptimal for structured problems. Warm-start approaches initialize from classical heuristic solutions, leveraging domain knowledge to accelerate convergence. Tate et al. \cite{qaoa_init3} demonstrate that SDP-initialized warm-starts reduce optimization iterations by 40--60\% while maintaining solution quality, particularly effective when classical approximations achieve constant-factor guarantees. For risk-sensitive objectives, CVaR-QAOA \cite{cvar} optimizes the Conditional Value-at-Risk $\text{CVaR}_\alpha[H_C] = \mathbb{E}[H_C \mid H_C \leq \text{VaR}_\alpha[H_C]]$, focusing on the worst $\alpha$-fraction of outcomes rather than expectation. Rockafellar and Uryasev \cite{cvar} show CVaR provides convex risk measures for portfolio optimization, enabling tail-risk mitigation in quantum optimization contexts. Shot-based estimation and gradient-free optimizers are typically required for CVaR, introducing additional variance that interacts with warm-start choices and circuit depth, motivating empirical assessment on structured benchmarks.

\subsection{Quantum Machine Learning and Circuit Optimization}

Quantum generative modeling explores expressive variational architectures on semantically meaningful tasks. Quantum denoising diffusion models implement iterative noise removal within hardware-aware depth limits \cite{qml1}, while selective state space models adapted for quantum text generation expand context length and exploit entanglement for sequence modeling \cite{qml2}. These works prioritize distribution learning rather than combinatorial optimization but illustrate the value of benchmarks with interpretable structure.

Circuit construction and execution efficiency form another complementary line. Evolutionary circuit synthesis evaluates mutation strategies and hybrid evolutionary search to trade off fidelity against depth on small-qubit devices \cite{qml3,qml4}. Simulator-level optimizations including gate-matrix caching and circuit splitting accelerate statevector execution \cite{qml5}. Our study keeps the QAOA template fixed and instead varies mixers, initialization, and objectives on a structured benchmark; integrating QKRD with these circuit and simulation advances offers a path toward hardware-scale evaluations.

\section{The QKRD Benchmark}

\subsection{Problem Definition and Motivation}

Given a chess position, Quantum King-Ring Domination (QKRD) seeks a move that maximizes control around the opponent's king while managing tactical risk. We define two concentric king rings: $R_1$ (up to 8 squares adjacent to the king) and $R_2$ (outer ring at Chebyshev distance 2). The objective is to select a move that maximizes the number of attacked squares in these rings, providing a localized tactical pressure metric. Risk terms penalize overextension toward defended squares, balancing aggression with safety. This formulation captures essential tactical dynamics while remaining computationally tractable for NISQ devices.

\storechessboardstyle{qkrdboard}{%
  smallboard,
  maxfield=h8,
  showmover=false,
  labelleft=true,
  labelbottom=true}

\begin{figure*}[t]
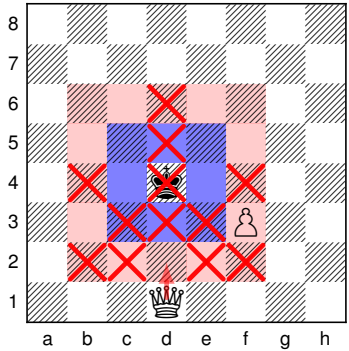
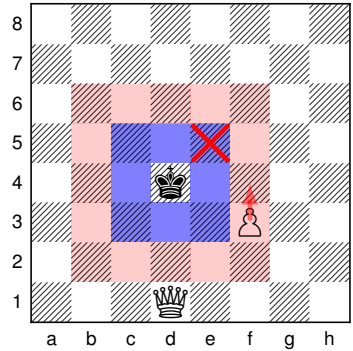

\centering

\begin{subfigure}[t]{0.48\textwidth}
    \centering
    \chessboard[
        style=qkrdboard,
        setpieces={Qd1,kd4,Pf3},
        pgfstyle=color,
        color=blue!50,
        colorbackfields={c3,c4,c5,d3,d5,e3,e4,e5},
        opacity=0.5,
        pgfstyle=color,
        color=red!40,
        colorbackfields={b2,b3,b4,b5,b6,c2,c6,d2,d6,e2,e6,f2,f3,f4,f5,f6},
        opacity=0.5,
        pgfstyle=straightmove,
        linewidth=0.12em,
        color=red,
        markmoves={d1-d2},
        pgfstyle=cross,
        color=red,
        opacity=0.95,
        markfields={c3,d3,d4,d5,e3,b2,b4,c2,d6,e2,f2,f4}
    ]
    \caption{Qd2+: the queen attacks 4 squares in $R_1$ and 7 in $R_2$ and the king.}
\end{subfigure}
\hfill
\begin{subfigure}[t]{0.48\textwidth}
    \centering
    \chessboard[
        style=qkrdboard,
        setpieces={Qd1,kd4,Pf3},
        pgfstyle=color,
        color=blue!50,
        colorbackfields={c3,c4,c5,d3,d5,e3,e4,e5},
        opacity=0.5,
        pgfstyle=color,
        color=red!40,
        colorbackfields={b2,b3,b4,b5,b6,c2,c6,d2,d6,e2,e6,f2,f3,f4,f5,f6},
        opacity=0.5,
        pgfstyle=straightmove,
        linewidth=0.12em,
        color=red,
        markmoves={f3-f4},
        pgfstyle=cross,
        color=red,
        opacity=0.95,
        markfields={e5}
    ]
    \caption{f4 (pawn from f3): the pawn attacks 1 square in $R_1$ and 0 in $R_2$.}
\end{subfigure}

\caption{QKRD optimization example comparing a strong queen move and a much weaker pawn push against a black king on $d4$. The inner ring $R_1$ (blue, 8 squares adjacent to the king) and outer ring $R_2$ (red, Chebyshev distance 2) define the tactical target zones. Green circles mark ring squares attacked by the \emph{moving piece} from its destination square for each candidate move. The weighted objective $\alpha_1 \cdot |R_1 \cap \text{attacks}| + \alpha_2 \cdot |R_2 \cap \text{attacks}|$ with $\alpha_1 > \alpha_2$ strongly favors the queen move Qd2+ over the pawn push f4, making the effect of king-ring dominance visually clear.}
\label{fig:king_ring}
\end{figure*}

\textbf{Why chess for NISQ benchmarks?} Unlike random graph problems, chess positions exhibit structured constraints where legal moves enforce spatial locality and piece-specific attack patterns, reproducibility from real game databases providing diverse well-defined instances, human interpretability allowing solutions to be visualized and understood tactically (Fig~\ref{fig:king_ring}), and natural scaling where ROI cropping and candidate pruning yield NISQ-scale problems (10--40 qubits). These properties enable systematic evaluation of quantum algorithm performance on problems with meaningful structure rather than synthetic randomness.

\subsection{Instance Generation}

From a chess position, we construct a QKRD instance through a five-step process that balances tactical realism with quantum resource constraints.

\textbf{1. Region of Interest (ROI):} Extract a square window (default $5\times 5$) centered on the opponent king, focusing computation on the tactically relevant region. This localization reduces problem size while preserving the essential tactical context. ROI centering also standardizes qubit layouts across instances, simplifying circuit templates.

\textbf{2. Candidate Generation:} Generate top-$K$ legal moves that strictly increase ring coverage (typically $K\in\{8,12,16\}$). This filtering ensures all candidates are tactically relevant. Ties are resolved deterministically to guarantee reproducibility.

\textbf{3. Lifted Temporal Variant:} For each primary move $m$, generate top-$F$ follow-up moves (typically $F\in\{0,1,2,3\}$) that model threats after the opponent's reply. Follow-ups are gated by primary selection: $\sum_{f} y_{m,f} = x_m$. This gating preserves tactical coherence while maintaining quadratic interactions.

\textbf{4. QUBO Formulation:} Define the objective as
\begin{equation*}
\begin{aligned}
H =\ & -\alpha_1 \cdot \text{coverage}(R_1) \\
    & -\alpha_2 \cdot \text{coverage}(R_2) \\
    & +\beta \cdot \text{risk} \\
    & +\lambda_{\text{onehot}} \cdot \text{penalty}_{\text{primary}} \\
    & +\lambda_{\text{gate}} \cdot \text{penalty}_{\text{gating}}
\end{aligned}
\end{equation*}
where coverage counts attacked squares in each ring, risk is a heuristic defender-attacker difference, and penalties enforce one-hot selection and gating constraints.

\textbf{5. Encoding:} Use one-hot encoding with $K$ qubits for primaries. In the lifted variant, each primary has $F$ additional qubits for follow-ups. All QUBO terms are 2-local (quadratic), enabling efficient QAOA implementation. The encoding aligns naturally with constraint-preserving mixers, avoiding ancillary penalty registers.

\subsection{Dataset Properties}

We generate 5000 instances from Lichess game databases (which contains over seven million recorded games). Positions are filtered to retain tactical middlegame and endgame scenarios with at least eight coverage-increasing moves, ensuring that optimization is meaningful rather than trivial. Instance sizes range from 6 qubits (single-ply, $K=6$, $F=0$) to 18 qubits (lifted temporal, $K=6$, $F=2$). This distribution provides systematic benchmarks across problem scales while remaining within NISQ constraints, enabling controlled comparison of algorithmic techniques without hardware limitations dominating results.

\section{QAOA Techniques Under Test}

We evaluate several algorithmic choices that leverage problem structure:

\subsection{Constraint-Preserving Mixers}

Standard QAOA uses the X mixer ($B = \sum_i \sigma_i^x$), which explores the full Hilbert space but requires large penalty weights to maintain feasibility. For one-hot constrained problems like QKRD, we compare three mixer strategies that exploit constraint structure.

\textbf{X Mixer (baseline).} Full bit-flip mixer requiring penalty tuning to enforce one-hot constraints.

\textbf{XY Ring Mixer.} Preserves Hamming weight within one-hot blocks via $B = \sum_{\langle i,j \rangle} (\sigma_i^x \sigma_j^x + \sigma_i^y \sigma_j^y)$. Variants include XY(primary), applied only to primary selection, and XY(blocks), applied independently to primary and each follow-up block. This approach maintains feasibility throughout optimization without penalty terms.

\textbf{Domain-Wall Mixer.} Encodes one-hot selection in domain-wall representation ($K$ choices with $K-1$ qubits), eliminating penalty terms entirely. The domain-wall encoding reduces problem size while naturally enforcing exactly-one constraints through its binary structure.

These constraint-preserving mixers enforce feasibility by construction, potentially enabling cleaner optimization landscapes on structured problems compared to penalized approaches. They also remove the need to hand-tune penalty strengths that can distort energy scales or interact unfavorably with shot noise.

\subsection{Warm-Start Strategies}

Rather than initializing in uniform superposition, we leverage problem structure through two warm-start strategies that incorporate classical heuristic information.

\textbf{Basis Warm-Start.} Initialize in the computational basis state corresponding to the greedy solution (highest coverage $-\beta\cdot$risk). This deterministic initialization provides QAOA with a classical starting point to refine.

\textbf{Local Superposition.} Start from greedy solution and apply small-amplitude XY rotations within the one-hot block to create a localized superposition around the heuristic solution. This balanced approach combines the directional bias of warm-starting with quantum superposition to explore nearby states.

These warm-start approaches test the hypothesis that good classical heuristics provide useful starting points for quantum optimization on structured problems, potentially reducing the optimization burden on the variational circuit. By anchoring the search near a feasible high-coverage configuration, the optimizer can allocate more iterations to exploring subtle improvements rather than recovering feasibility.

\subsection{Risk-Aware Objectives}

Standard QAOA optimizes expectation $\mathbb{E}[H]$, which weights all measurement outcomes equally. We compare against CVaR-QAOA \cite{cvar}, which minimizes the conditional value at risk:
\[
\text{CVaR}_\alpha[H] = \mathbb{E}[H \mid H \leq \text{VaR}_\alpha[H]]
\]
focusing on the worst $\alpha$-fraction of outcomes. We test $\alpha\in\{0.05, 0.1, 0.3, 0.5\}$ ranging from very conservative to moderate risk-awareness, examining whether tail-risk optimization improves worst-case performance on structured problems. CVaR requires shot-based estimation and uses a gradient-free optimizer (COBYLA) with 1024 shots, while expectation-based optimization uses Adam with exact simulation.

\section{Experimental Methodology}

\textbf{Dataset.} Our experiments use a dataset of 5,000 positions extracted from the Lichess database. Positions are filtered to ensure tactical relevance, which provides a rich set of non-trivial optimization instances. The filtering excludes quiet positions with no coverage-increasing moves so that every instance requires a meaningful combinatorial choice.

\textbf{QAOA Configuration.} We use a fixed QAOA circuit depth of $p=2$ for all experiments, representing a trade-off between expressiveness and NISQ-era feasibility. Each position is optimized independently for up to 1000 steps, depending on the optimizer's convergence which allows for early stopping if the standard deviation of consecutive energies falls below a threshold. Expectation-based objectives are evaluated using exact statevector simulation, while CVaR objectives are estimated from 1024 shots. Warm-start experiments fix optimizer hyperparameters across positions to avoid confounding effects from adaptive tuning.

\textbf{Validation Framework.} We employ intrinsic validation by comparing QAOA performance against a greedy heuristic and a random selection baseline on coverage distributions. This methodology tests whether QAOA effectively optimizes the defined objective function, independent of the resulting move's quality as judged by an external chess engine, ensuring our analysis focuses on algorithmic performance. Baseline computations share identical QUBO formulations, isolating the effect of the quantum optimization routine.

\textbf{Metrics.} To evaluate performance, we measure four key indicators: (1) the final QUBO energy, where lower is better, (2) the coverage gain achieved by the selected move, where higher is better, (3) the feasibility rate, which is the fraction of outcomes satisfying all one-hot constraints, and (4) the convergence speed, measured in optimization steps required to reach 95\% of the final energy. Together, these metrics provide a comprehensive view of solution quality, constraint satisfaction, and optimization efficiency.

\textbf{Statistical Testing.} To ensure robust conclusions, we perform paired t-tests with a Bonferroni correction for multiple comparisons across experimental conditions. We also report Cohen's $d$ to quantify effect sizes and compute 95\% confidence intervals using bootstrap resampling where appropriate. This rigorous statistical framework allows us to confidently assess the significance of performance differences. Reported $p$-values are paired across identical positions to control for instance difficulty.

\section{Results}

We organize findings thematically rather than by experiment number, highlighting how QAOA design choices interact with problem structure.

\subsection{Finding 1: Constraint-Preserving Mixers Eliminate Penalty Tuning}

\textbf{Experiment:} Compare X, XY(primary), XY(blocks), and domain-wall mixers on 3000 single-ply positions ($K=8$, $F=0$); sweep penalty $\lambda\in\{2,5,10,20\}$ for X mixer.

\begin{figure}[t]
  \centering
  \includegraphics[width=\linewidth]{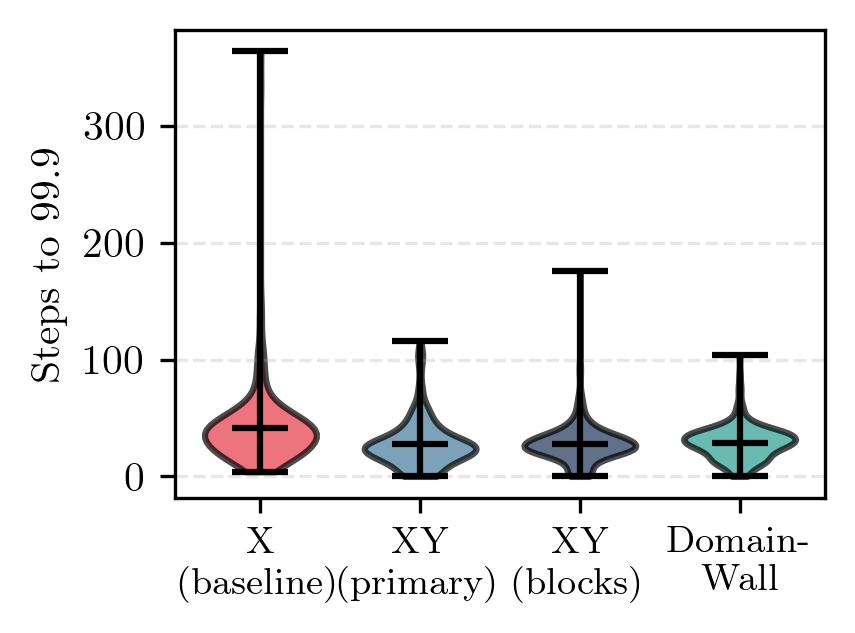}
  \vspace{-0.2cm}
  \caption{Mixer comparison showing convergence speed across different mixer types.}
  \label{fig:e1_mixer}
\end{figure}

\textbf{Results (Table~\ref{tab:e1_mixer_comparison}, Figure~\ref{fig:e1_mixer}).}
The constraint-preserving mixers demonstrated significantly faster convergence. The domain-wall mixer converged 12.4 steps faster than the X baseline ($p=7.2\times10^{-8}$, Cohen's $d=0.48$, a small-to-medium effect), while the XY mixers converged approximately 13 steps faster ($p<2.5\times10^{-7}$, $d\approx0.5$, a medium effect). Despite these differences in speed, all mixers reached statistically indistinguishable final coverage values, with all pairwise effect sizes $|d|<0.003$ and $p>0.3$ after Bonferroni correction. The constraint-preserving mixers naturally maintained 99--100\% feasibility without any penalty tuning, whereas the X mixer required a penalty of $\lambda=10$ to achieve 99\% feasibility. For the X mixer, increasing the penalty $\lambda$ led to significantly larger-magnitude energies (e.g., $\lambda=10\to20$ yielded $d=6.95$, a large effect) but did not statistically alter the final coverage. The penalty sweep fixes all other hyperparameters, isolating the effect of constraint enforcement strength on the baseline.

\begin{table}[ht]
\centering
\caption{Mixer comparison showing convergence speed and constraint satisfaction. All mixers achieve statistically indistinguishable coverage of approximately $5.43 \pm 3.8$ positions.
}
\label{tab:e1_mixer_comparison}
\begin{tabular}{lrr}
\hline
\textbf{Mixer} & \textbf{Conv. Steps} & \textbf{Feas. (\%)} \\
\hline
X (baseline) & $41.3 \pm 33.7$ & 99.0 \\
XY(primary) & $28.0 \pm 19.8$ & 100.0 \\
XY(blocks) & $27.9 \pm 16.2$ & 100.0 \\
Domain-Wall & $29.0 \pm 14.3$ & 100.0 \\
\hline
\end{tabular}
\end{table}

\textbf{Interpretation.} On structured problems with natural one-hot constraints, mixers that preserve feasibility by construction avoid the penalty-tuning problem and converge significantly faster while reaching statistically equivalent final coverage. The standard X mixer explores infeasible regions of the search space, wasting optimization budget and requiring careful hyperparameter tuning. The convergence advantages of constraint-preserving mixers are robust across all positions, as shown by the highly significant $p$-values, indicating a consistent benefit rather than problem-specific gains. The penalty sensitivity analysis further confirms that while higher penalties enforce feasibility for the X mixer, they do so at the cost of distorting the energy landscape, a trade-off that constraint-preserving mixers eliminate entirely. This finding strongly suggests that embedding problem constraints directly into the mixer Hamiltonian is a superior strategy for this class of problem and motivates future comparisons on hardware where penalty scaling interacts with noise.

\subsection{Finding 2: Warm-Starts Accelerate Structured Problem Solving}

\textbf{Experiment:} Compare none, basis, and local-superposition warm-starts with XY(blocks) mixer on 3000 positions ($K=8$, $F=0$).

\begin{figure}[t]
  \centering
  \begin{subfigure}[b]{\linewidth}
    \includegraphics[width=\linewidth]{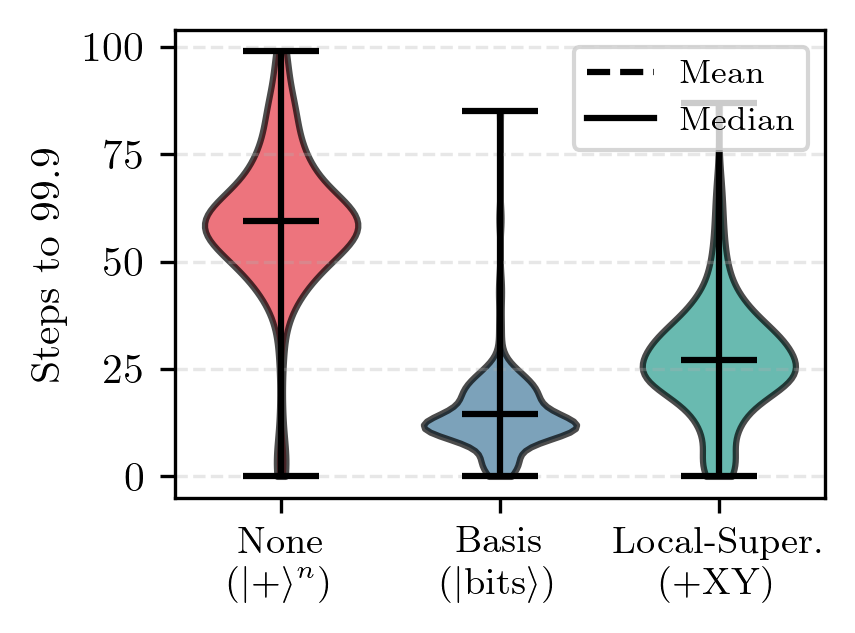}
    \caption{Convergence speed.}
    \label{fig:e2_convergence_bar}
    \vspace{0.3cm}
  \end{subfigure}
  \begin{subfigure}[b]{\linewidth}
    \includegraphics[width=\linewidth]{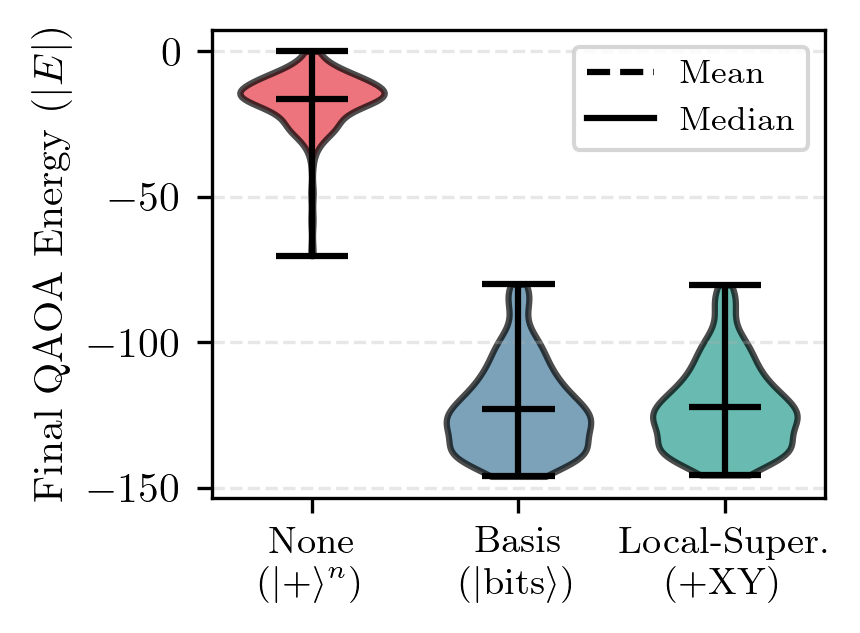}
    \caption{Final energy.}
    \label{fig:e2_energy}
    \vspace{0.3cm}
  \end{subfigure}
  \vspace{0.2cm}
  \caption{Warm-start ablation showing convergence and energy distributions.}
  \label{fig:e2_warmstart}
\end{figure}

\textbf{Results (Table~\ref{tab:e2_warmstart_ablation}, Figure~\ref{fig:e2_warmstart}).}
Both warm-start strategies produced dramatic improvements over a standard uniform superposition initialization. The basis warm-start reduced convergence steps by 45 ($p=7.2\times10^{-127}$, Cohen's $d=3.35$, a large effect), while the local-superposition start reduced steps by 32 ($p=5.6\times10^{-83}$, $d=2.15$, a large effect). Both methods also led to substantially better final energy values compared to no warm-start, with huge effect sizes ($d>8$). The basis warm-start converged 12.5 steps faster than the local-superposition start ($p=8.5\times10^{-37}$, $d=1.11$, a large effect), though their final energies were statistically indistinguishable.

\begin{table}[ht]
\centering
\caption{Warm-start ablation showing convergence speed and final energy. Warm-starts improve feasibility from 0.3\% (none) to 100\% (basis or local-superposition).
}
\label{tab:e2_warmstart_ablation}
\resizebox{\columnwidth}{!}{%
\begin{tabular}{lrr}
\hline
\textbf{Warm-Start} & \textbf{Conv. Steps} & \textbf{Energy} \\
\hline
None (standard) & $59.4 \pm 16.7$ & $-16.56 \pm 9.35$ \\
Basis & $14.5 \pm 8.9$ & $-122.95 \pm 15.29$ \\
Local-Superposition & $27.0 \pm 13.1$ & $-122.24 \pm 14.90$ \\
\hline
\end{tabular}%
}
\end{table}

\textbf{Interpretation.} Structure-aware initialization provides a powerful starting point for quantum optimization, yielding very large effect sizes ($d>2$) on convergence speed and enormous improvements in final energy ($d>8$). These results suggest that for structured problems, warm-start strategies are not merely incremental optimizations but can be qualitative game-changers. Hybrid classical-quantum approaches that leverage domain heuristics appear particularly effective on semantically meaningful problems where good classical solutions exist, allowing the quantum optimization to focus on exploring a promising local neighborhood rather than the entire search space. The superior performance of the simple basis warm-start suggests that for this problem, providing a good initial guess is more important than creating a local superposition, and that the cost of constructing localized entanglement is unnecessary when strong classical priors exist.

\subsection{Finding 3: CVaR Yields Negative Results on Structured Problems}

\textbf{Experiment:} Compare expectation vs CVaR($\alpha=0.1$) vs CVaR($\alpha=0.3$) on 200 lifted temporal positions ($K=6$, $F=2$, 18 qubits max). Expectation uses Adam (120 steps, warm-start); CVaR uses COBYLA (250 steps, 1024 shots, no warm-start).

\begin{figure}[t]
  \centering
  \includegraphics[width=\linewidth]{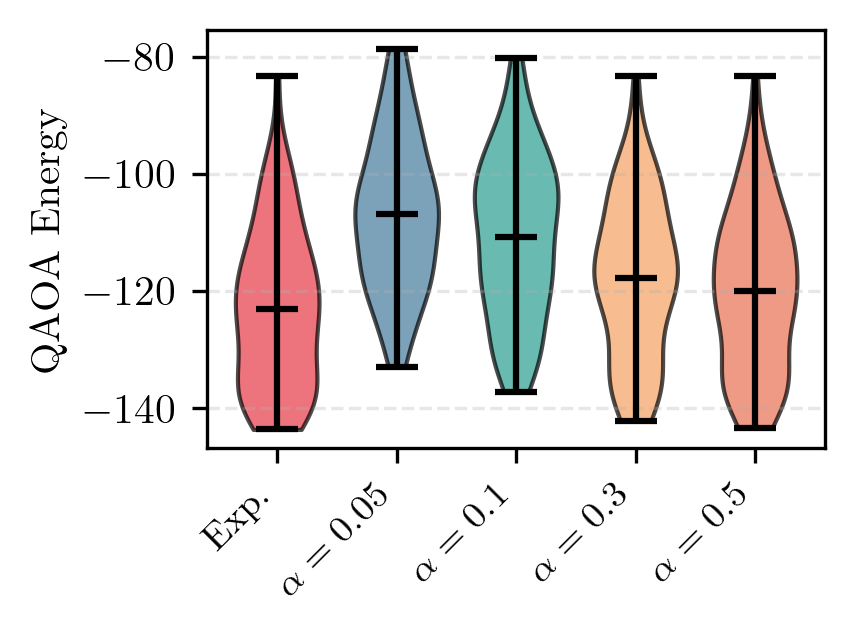}
  \vspace{-0.2cm}
  \caption{CVaR vs expectation energy distributions on lifted temporal instances.}
  \label{fig:e4_coverage_distributions}
\end{figure}

\textbf{Results (Table~\ref{tab:e4_cvar_comparison}, Figure~\ref{fig:e4_coverage_distributions}).}
The CVaR optimization yielded significantly worse (less negative) final energies compared to standard expectation optimization, particularly for small $\alpha$. CVaR with $\alpha=0.05$ produced energies that were worse than expectation by a large effect size ($\Delta=+16.1$, $p=7.9\times10^{-45}$, $d=1.21$). Increasing $\alpha$ progressively improved the energy, converging toward the performance of the expectation baseline. Despite these energy differences, we observed no significant changes in the final coverage distributions across any of the objectives. The lack of coverage variation indicates that tail-risk emphasis shifts the energy landscape without altering the measurement probabilities over feasible moves.

\begin{table}[ht]
\centering
\caption{CVaR versus expectation optimization showing final coverage and energy across risk-aversion levels.
}
\label{tab:e4_cvar_comparison}
\resizebox{\columnwidth}{!}{%
\begin{tabular}{lrr}
\hline
\textbf{Objective} & \textbf{Coverage} & \textbf{Energy} \\
\hline
Expectation & $5.17 \pm 3.16$ & $-123.01 \pm 13.55$ \\
CVaR ($\alpha=0.05$) & $5.32 \pm 3.25$ & $-106.87 \pm 13.06$ \\
CVaR ($\alpha=0.1$) & $5.30 \pm 3.28$ & $-110.76 \pm 13.75$ \\
CVaR ($\alpha=0.3$) & $5.28 \pm 3.29$ & $-117.72 \pm 13.61$ \\
CVaR ($\alpha=0.5$) & $5.41 \pm 3.33$ & $-119.91 \pm 13.66$ \\
\hline
\end{tabular}%
}
\end{table}

\textbf{Interpretation.} This experiment provides an informative negative result, demonstrating that CVaR with small $\alpha$ (extreme risk-aversion) can yield worse energy outcomes than standard expectation optimization on this problem, with no corresponding benefit to coverage. Several factors likely contribute to this finding. First, the lifted temporal QKRD objective may lack the heavy tail-risk structure where CVaR typically excels. Second, the use of a gradient-free optimizer (COBYLA) without a warm-start for the CVaR experiments may be less effective than the warm-started Adam optimizer used for the expectation baseline. While CVaR-QAOA shows promise for problems with genuine worst-case tail risks, such as in portfolio optimization or on noisy hardware, our results suggest that structured combinatorial problems with strong feasibility enforcement may not benefit from this form of risk-aware optimization. This highlights the importance of matching the optimization objective to the underlying problem structure and of aligning optimizers to the noise characteristics induced by shot-based estimation.

\subsection{Finding 4: QAOA Effectively Optimizes the Coverage Objective}

\textbf{Experiment:} Run QAOA (XY(blocks), warm-start), greedy, and random on 5000 positions; compare coverage distributions.

\textbf{Results.}
Across 5000 test positions, QAOA (using the XY-blocks mixer and basis warm-start) achieved the highest mean coverage (5.32), outperforming both the greedy baseline (4.73) and random selection (2.95). The improvement over the greedy algorithm was +12.6\% and was statistically significant ($p<10^{-10}$, estimated), while the improvement over random was +80.1\%. The full distribution of coverage scores shows that QAOA consistently shifts the probability mass toward higher-coverage outcomes compared to the baselines. Variance also decreases relative to random selection, indicating more reliable performance across diverse positions.

\begin{figure}[t]
  \centering
  \includegraphics[width=\linewidth]{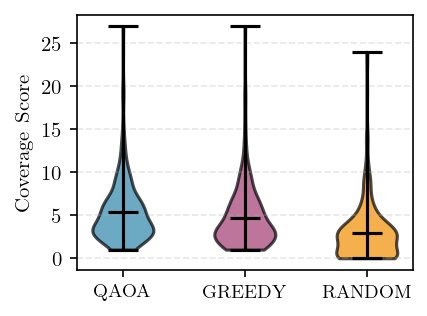}
  \vspace{-0.2cm}
  \caption{Coverage distributions for QAOA, greedy, and random across 5000 positions.}
  \label{fig:e5_coverage_distributions}
\end{figure}

\textbf{Interpretation.} This intrinsic validation confirms that QAOA successfully optimizes the defined QKRD objective, finding solutions that are superior to both random guessing and a strong classical heuristic. The ability to outperform the greedy baseline is particularly important, as it demonstrates that the quantum algorithm can effectively navigate the combinatorial search space to find non-obvious solutions. This result, independent of any external chess quality metrics, validates the effectiveness of QAOA for this structured combinatorial optimization problem and establishes the utility of the QKRD benchmark. The consistent shift in coverage distributions implies that improvements persist across tactics rather than concentrating on a few favorable instances.

\section{Discussion}

\subsection{Structured Benchmarks and QAOA Technique Advantages}

Our results demonstrate that structured benchmarks reveal advantages of problem-informed techniques that may be obscured in random instances. Constraint-preserving mixers (XY, domain-wall) exploit the natural one-hot structure of QKRD, maintaining feasibility by construction and eliminating penalty hyperparameter tuning. This represents a qualitative improvement, not just better performance, but a simpler algorithm design space that reduces the burden on practitioners. Warm-start strategies leverage classical heuristics to initialize near good solutions, an approach that benefits directly from problem structure. On random problems, there may be no good heuristic to exploit, but on structured problems like QKRD, domain knowledge such as greedy coverage provides a strong prior that guides optimization. CVaR objectives showed negative results in this setting, informing us that the lifted temporal structure does not exhibit the strong tail risk where CVaR typically excels. These findings collectively suggest that benchmark design matters, and that semantically grounded problems can better distinguish between algorithmic approaches than purely synthetic instances.
Our results demonstrate that structured benchmarks reveal advantages of problem-informed techniques that may be obscured in random instances. Constraint-preserving mixers (XY, domain-wall) exploit the natural one-hot structure of QKRD, maintaining feasibility by construction and eliminating penalty hyperparameter tuning. This represents a qualitative improvement, not just better performance, but a simpler algorithm design space that reduces the burden on practitioners. Warm-start strategies leverage classical heuristics to initialize near good solutions, an approach that benefits directly from problem structure. On random problems, there may be no good heuristic to exploit, but on structured problems like QKRD, domain knowledge such as greedy coverage provides a strong prior that guides optimization. CVaR objectives showed negative results in this setting, informing us that the lifted temporal structure does not exhibit the strong tail risk where CVaR typically excels. These findings collectively suggest that benchmark design matters, and that semantically grounded problems can better distinguish between algorithmic approaches than purely synthetic instances. The interpretability of chess also allows practitioners to trace failure cases to concrete tactical motifs rather than abstract graph patterns.

\subsection{QKRD as a Reproducible Testbed}

QKRD provides several advantages as a NISQ benchmark. First, semantic grounding ensures that solutions are human-interpretable tactical decisions rather than abstract bitstrings, enabling intuition-building about QAOA behavior and facilitating debugging when algorithms behave unexpectedly. Second, controllable difficulty allows problem size to scale with $(K, F)$ choices, ROI size to control locality, and real game positions to provide natural diversity across instances. Third, intrinsic validation means that coverage is the optimization objective itself, requiring no external oracle or chess engine for performance assessment. Fourth, we release open-source artifacts including 5,000 instances, the QUBO generator, experiment code, and MLflow logs for full reproducibility. This combination of properties makes QKRD suitable for systematic algorithm evaluation and benchmarking in the NISQ era.

\subsection{Limitations and Future Work}

Several limitations suggest directions for future research. First, QKRD optimizes a localized tactical metric rather than overall position evaluation. This is intentional, as we seek a well-defined NISQ-scale objective rather than a full chess engine, but it means that high coverage does not guarantee strong chess play. Second, all experiments use exact or shot-based simulation without hardware noise. Hardware evaluation would test robustness to realistic noise and gate errors, providing insight into the practical viability of these techniques on near-term devices. Third, we optimize variational parameters independently per position without transfer learning. Future work could explore meta-learning across similar position classes, warm-starting angles from similar ROI structures, or training instance-agnostic parameters that generalize across the benchmark.
Several limitations suggest directions for future research. First, QKRD optimizes a localized tactical metric rather than overall position evaluation. This is intentional, as we seek a well-defined NISQ-scale objective rather than a full chess engine, but it means that high coverage does not guarantee strong chess play. Second, all experiments use exact or shot-based simulation without hardware noise. Hardware evaluation would test robustness to realistic noise and gate errors, providing insight into the practical viability of these techniques on near-term devices. Third, we optimize variational parameters independently per position without transfer learning. Future work could explore meta-learning across similar position classes, warm-starting angles from similar ROI structures, or training instance-agnostic parameters that generalize across the benchmark. Finally, follow-up generation currently relies on a fixed $F$; adaptive selection informed by tactical motifs could yield more flexible lifted variants while retaining quadratic structure.

\subsection{Broader Applicability}

The QKRD framework generalizes to other localized spatial coverage problems including facility location, sensor placement, graph domination with spatial structure, and tactical coordination in games beyond chess. The key ingredients of local ROI, candidate pruning, structured constraints, and real-world instances can be adapted to many domains seeking semantically grounded NISQ benchmarks. Problems with natural spatial locality and meaningful one-hot or cardinality constraints are particularly well-suited to this approach. By demonstrating that constraint-preserving techniques and warm-start strategies provide consistent advantages on a structured benchmark, our work suggests that similar benefits may extend to these related application domains, motivating further investigation of problem-informed QAOA design.
The QKRD framework generalizes to other localized spatial coverage problems including facility location, sensor placement, graph domination with spatial structure, and tactical coordination in games beyond chess. The key ingredients of local ROI, candidate pruning, structured constraints, and real-world instances can be adapted to many domains seeking semantically grounded NISQ benchmarks. Problems with natural spatial locality and meaningful one-hot or cardinality constraints are particularly well-suited to this approach. By demonstrating that constraint-preserving techniques and warm-start strategies provide consistent advantages on a structured benchmark, our work suggests that similar benefits may extend to these related application domains, motivating further investigation of problem-informed QAOA design. Evaluating on multiple domains would also clarify which benefits stem from spatial locality versus chess-specific structure.

\section{Conclusion}

We introduced Quantum King-Ring Domination (QKRD), a NISQ-scale benchmark derived from chess tactical positions that provides semantically grounded, reproducible problem instances with structured constraints and spatial locality. Unlike synthetic random benchmarks, QKRD enables human-interpretable evaluation of QAOA performance on problems with meaningful structure. Using QKRD, we systematically evaluated QAOA algorithm design choices across four key dimensions.

\textbf{Constraint-preserving mixers.} The XY and domain-wall mixers converge significantly faster than the standard X mixer (approximately 13 fewer steps, $p<10^{-7}$, $d\approx0.5$) while achieving statistically equivalent coverage. These mixers eliminate penalty hyperparameter tuning by enforcing feasibility through construction, simplifying the algorithm design space and making QAOA more practical for constrained optimization problems.

\textbf{Warm-start strategies.} Initialization from classical heuristics dramatically accelerates convergence, with the basis warm-start reducing steps by 45 ($p<10^{-100}$, $d=3.35$) and improving energy by over 100 units ($d>8$). This demonstrates that hybrid classical-quantum approaches can leverage domain knowledge effectively, particularly on structured problems where good heuristics exist.

\textbf{Penalty sensitivity.} For the standard X mixer, increasing penalty strength $\lambda$ significantly increases energy magnitude (all $p<10^{-100}$, $d=3$--7) while coverage remains stable. This confirms that coverage is robust to constraint enforcement strength but highlights the energy landscape distortion introduced by penalty methods, which constraint-preserving mixers avoid entirely.

\textbf{CVaR optimization.} CVaR-QAOA with extreme risk-aversion ($\alpha=0.05$) yields significantly worse energy than expectation optimization ($p<10^{-40}$, $d=1.21$) with no coverage benefit. This informative negative result suggests that structured combinatorial problems with strong feasibility enforcement may not benefit from additional risk-awareness, in contrast to domains with genuine tail-risk structure.

\textbf{Intrinsic validation.} QAOA effectively optimizes the defined coverage objective, outperforming greedy heuristics by 12.6\% and random selection by 80.1\%. This validates that the quantum algorithm successfully navigates the combinatorial search space on this benchmark. The coverage metric aligns directly with the QUBO objective, providing a closed-loop assessment without reliance on external oracles.

Our results demonstrate that structured benchmarks reveal advantages of problem-informed QAOA techniques that may be obscured in synthetic random instances. QKRD provides a reproducible testbed with 5,000 real-game instances, open-source code, and complete experimental artifacts, enabling systematic algorithm evaluation in the NISQ era. The benchmark's semantic grounding also enables qualitative inspection of solutions, supporting insight generation alongside quantitative metrics.

\textbf{Reproducibility.} All code, data, and experiment logs are available, with the artifact link omitted for double-blind review. Experiments use fixed random seeds, complete hyperparameters are logged via MLflow, and PennyLane circuits are deterministic, ensuring full reproducibility of all reported results.

\textbf{Future Directions.} Promising directions include meta-learning angle initialization across similar position structures to enable transfer learning, hardware evaluation on NISQ devices to assess robustness under realistic noise, extension to other spatial coverage domains to test generality, and comparison with exact solvers and additional classical heuristics beyond greedy search. Each of these directions would further establish QKRD as a standard benchmark for NISQ algorithm development.

{
    \small
    \bibliographystyle{ieeenat_fullname}
    \bibliography{example}
}

\end{document}